\pdfoutput=1

\documentclass[11pt]{article}

\usepackage[]{acl}

\usepackage{times}
\usepackage{latexsym}

\usepackage[T1]{fontenc}

\usepackage[utf8]{inputenc}

\usepackage{microtype}

\usepackage{inconsolata}

\usepackage{latexsym}
\usepackage{graphicx}
\usepackage{multirow}
\usepackage{rotating}
\usepackage{amsmath} 
\usepackage{subfig}
\usepackage{colortbl}
\usepackage{booktabs}
\usepackage{indentfirst}
\usepackage{adjustbox}
\usepackage{natbib}
\usepackage{hyperref}
\hypersetup{
    colorlinks=true,    
    linkcolor=[rgb]{0,0,0.6}, 
    filecolor=[rgb]{0,0,0.6}, 
    urlcolor=[rgb]{0,0,0.6},  
    citecolor=[rgb]{0,0,0.6}  
}

\title{Contrastive and Consistency Learning for Neural Noisy-Channel Model in Spoken Language Understanding}

\author{Suyoung Kim, Jiyeon Hwang, Ho-Young Jung \\
  Department of AI, Kyungpook National University\\
  \texttt{\{swimai, exo1926, hoyjung\}@knu.ac.kr} 
  }

\begin{document}
\maketitle

\begin{abstract}
Recently, deep end-to-end learning has been studied for intent classification in Spoken Language Understanding (SLU). However, end-to-end models require a large amount of speech data with intent labels, and highly optimized models are generally sensitive to the inconsistency between the training and evaluation conditions. Therefore, a natural language understanding approach based on Automatic Speech Recognition (ASR) remains attractive because it can utilize a pre-trained general language model and adapt to the mismatch of the speech input environment. Using this module-based approach, we improve a noisy-channel model to handle transcription inconsistencies caused by ASR errors. We propose a two-stage method, \textit{Contrastive and Consistency Learning (CCL)}, that correlates error patterns between clean and noisy ASR transcripts and emphasizes the consistency of the latent features of the two transcripts. Experiments on four benchmark datasets show that CCL outperforms existing methods and improves the ASR robustness in various noisy environments. Code is available at \hyperlink{https://github.com/syoung7388/CCL}{https://github.com/syoung7388/CCL}
\end{abstract}

\section{Introduction}
\label{Introduction}
Understanding a conversation is critical for reaching a user’s goal in a spoken dialogue system. Spoken Language Understanding (SLU) has been investigated by applying an end-to-end approach with a singular speech input model for training. The end-to-end approach requires an extensive volume of labeled speech data, which can be challenging to collect. As dialogue systems are commonly used in many real-world scenarios, the end-to-end SLU models are susceptible to unseen instances that mismatch the training data due to limited language expression and acoustic conditions. Utilizing a module-based approach, which leverages the capabilities of pre-trained language models, is an appealing option for developing general spoken dialogue systems. However, this option depends on the premise that the model pre-trained on clean text can effectively handle errors caused by the Automatic Speech Recognition (ASR) module. Consequently, it is necessary to solve the problem of inconsistencies in noisy ASR transcripts that contain speech recognition errors.

Recent studies have addressed the challenge of discrepancies between clean and noisy ASR transcripts. \citet{ruan2020towards} minimized the Kullback-Leibler (KL) divergence so that the prediction distributions of both transcripts become similar in a noisy environment. In contrast, \citet {chang22c_interspeech} demonstrated a technique that utilizes a contrastive learning to align the implicit features of paired clean and noisy ASR transcripts. However, most previous studies still need to consider the alignment for fine-grained errors such as insertion, deletion, and substitution caused by ASR modules.

We expect that using an improved noisy-channel model will mitigate the sensitivity of the pre-trained language model to discrepancies between clean and noisy ASR transcripts. The noisy-channel approach aims to identify the target word, even if the input becomes scrambled or distorted. We present the neural noisy-channel model to predict error-free intent by (i) finding the error portions of given noisy input relative to clean transcript and (ii) matching the latent representations from two transcripts. In this paper, we introduce \textit{Contrastive and Consistency Learning (CCL)} to perform this two-stage.

\begin{figure*}[t]
    \centering
    \includegraphics[width=16cm]{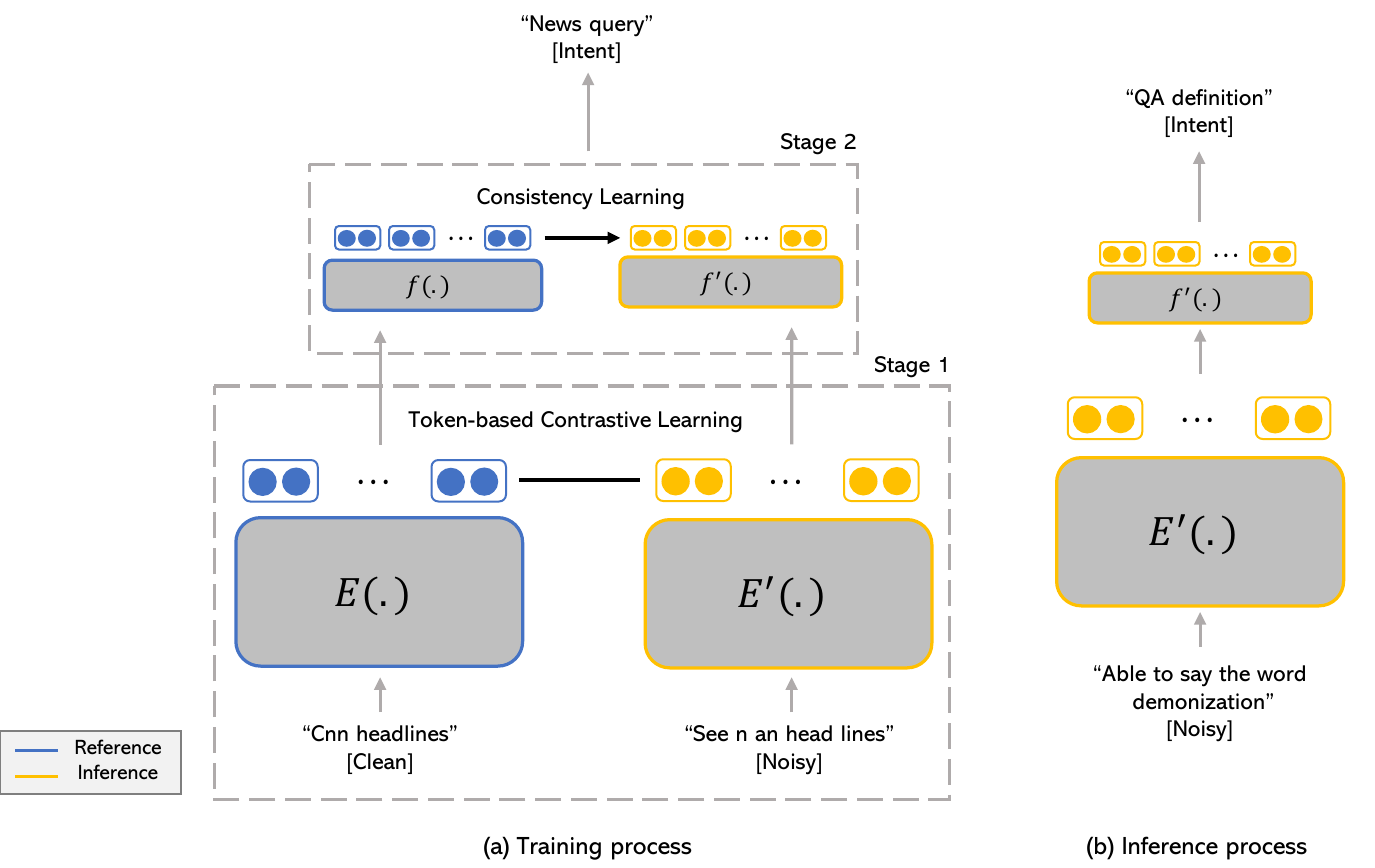}
    \caption{An overview of CCL. (a) Given a noisy ASR transcript, the inference network correlates error portions through token-based contrastive learning and then maintains coherence with the latent feature of the clean transcript through consistency learning. (b) In evaluation, inference network takes only noisy ASR transcript as input.}
    \label{figure1}
\end{figure*}

As shown in Figure \ref{figure1} (a), reference and inference networks receive clean and noisy ASR transcripts as inputs, respectively. During the CCL, two networks simultaneously perform a token-based contrastive learning followed by a consistency learning. The purpose of token-based contrastive learning is to correlate errors in the noisy ASR transcript with the corresponding clean transcript at both word and utterance token-levels. We select the word tokens as a positive pair by aligning them with an edit distance algorithm when two transcriptions do not match sequentially. In addition, consistency learning makes the inference network emphasize the coherence between clean and noisy latent features to avoid misclassifying the noisy ASR transcriptions. In Figure \ref{figure1} (b), we only use the inference network during the evaluation process, and the reference network plays a supplementary role in inference network training. This allows us to solve the ASR error problem without any increase in inference time. 

To demonstrate the effectiveness of the proposed CCL, we experiment with the SLU benchmark datasets: SLURP \citep{bastianelli-etal-2020-slurp}, Timers  \citep{lugosch2021timers}, FSC \citep{lugosch2019speech}, and SNIPS \citep{coucke2018snips}. On the SLURP benchmark dataset, the proposed CCL method improves the intent classification performance by 2.59\% under severe ASR errors. We also observe that our method outperforms previous module-based or end-to-end approaches on SNIPS. To summarize, our contributions are as follows:
\begin{itemize}
\item We investigate the error rates and the error portion that occurs during speech recognition.
\item We propose two-stage CCL method to enhance the noisy-channel model. This method correlates the error parts between two transcripts and emphasizes the consistency of latent features. 
\item We demonstrate that the CCL method improves the performance for noisy ASR transcripts through extensive experiments on various benchmark datasets. In addition, we study the robustness of ASR errors by conducting ablation studies and visualizing the representations to investigate the effectiveness of our proposed method.
\end{itemize}

\section{Related Work}
\label{Related Work}
In this section, we introduce the related work from two aspects.
\subsection{Spoken Language Understanding}
SLU can be implemented by two main approaches: (i) a module-based approach, wherein speech is converted to text through ASR followed by natural language understanding, and (ii) an end-to-end approach, wherein the system is a single network from input to task performance. In the module-based approach, attention-based SLU studies  \citep{liu2016attention, goo2018slot} achieved good performances with the introduction of the RNN-based attention technique \citep{bahdanau2015neural}. \citet{liu2016attention} used an attention-based bidirectional LSTM structure for simultaneous intent classification and slot filling. In a follow-up study, \citet{goo2018slot} added slot gates to BLSTM to correlate intents and slots. In contrast, the end-to-end approach has the advantage of being managed as a single network. In end-to-end approaches, a combined CNN and LSTM structure \citep{chen2018spoken, cao2021sequential} and a GRU-only structure \cite{serdyuk2018towards} were proposed. Although the studies \citep{seo2022integration, saxon21_interspeech, qian2021speech} employing the Transformer architecture have demonstrated promising performance, there are limitations in training with large-scale data to develop a general dialogue system.

\begin{table}[t]
\setlength{\tabcolsep}{10pt}
\renewcommand{\arraystretch}{1.5}
\resizebox{\columnwidth}{!}{
\begin{tabular}{c|cccc}
\toprule[1.2pt]
\textbf{Dataset} & \textbf{SLURP} & \textbf{Timers} & \textbf{FSC} & \textbf{SNIPS} \\ \midrule
Google API  & 0.24 & 0.15 & 0.08 & - \\
HuBERT  & 0.44 & 0.86 & 0.17 & - \\
Wav2vec2.0  & 0.56 & 0.57 & 0.27 & 0.41 \\ \bottomrule[1.2pt]
\end{tabular}
}
\caption{Word Error Rate (WER) of noisy ASR transcripts on the benchmark datasets.}
\label{table_1}
\end{table}

\subsection{ASR Errors in a Module-based Approach} 
By utilizing the pre-trained language models, such as BERT \citep{devlin2018bert} and RoBERTa \citep{liu2019roberta}, numerous studies in the field of SLU have successfully enhanced their generalization capabilities. \citet{chen2019bert} combined intent classification and slot filling with the output of BERT for clean transcripts and achieved better performance than existing attention-based models. \citet{qin2021co} proposed the BERT structure with the addition of a co-interactive module that considered slot and intent relationships. However, their exclusive training on large-scale clean datasets made the pre-trained language model susceptible to ASR errors. \citet {ruan2020towards} endeavored to mitigate these recognition errors by utilizing the KL divergence between the predicted probabilities for clean and noisy ASR transcripts. In contrast, \citet{kim2022improved} used the combined cross-entropy loss of two transcripts to predict intent. \citet{chang22c_interspeech} used contrastive learning to learn the error relationship between two transcripts during additional pre-training, and the noisy problem was mitigated by supervised contrastive learning and self-distillation during the fine-tuning.

\section{Method}
\label{Method}

\begin{figure*}[t]
    \centering
    \includegraphics[width=\textwidth]{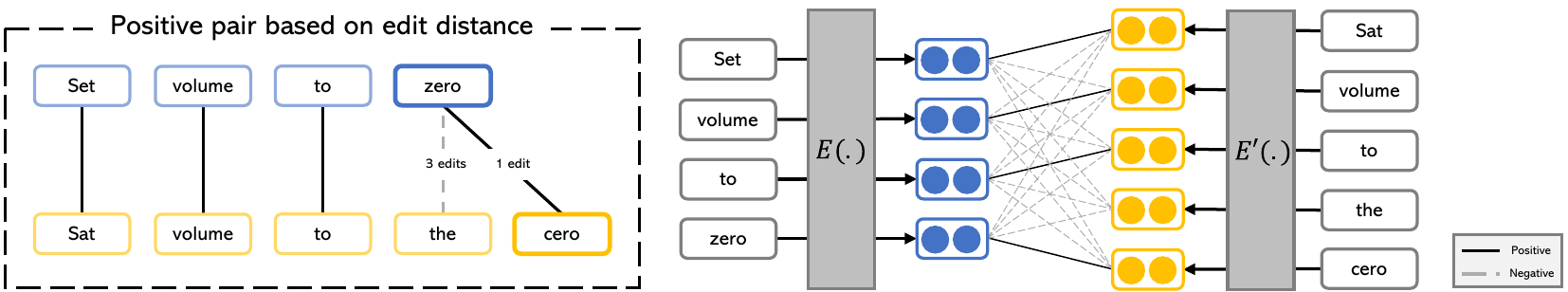}
    \caption{Selective-token contrastive learning finds a positive pair based on the edit distance for the given clean transcript “\texttt{Set volume to zero}” and noisy ASR transcript “\texttt{Sat volume to the cero}”, and conducts contrastive learning as shown on the right side.}
    \label{figure2}
\end{figure*}
\begin{table}[t]
\renewcommand{\arraystretch}{1.2}
\resizebox{\columnwidth}{!}{
\begin{tabular}{c|c|c|c} \toprule[1.2pt]
\textbf{Clean transcript}                                                                                             & \textbf{Noisy ASR transcript}                                                                                         & \textbf{WER} & \textbf{ASR errors}                                            \\ \midrule
“turn up”                                                                                               & “turn off”                                                                                          & 0.5          & up → off                                                              \\ \midrule
\begin{tabular}[c]{@{}c@{}}“brighten the lights \\ a little bit”\end{tabular}                           & \begin{tabular}[c]{@{}c@{}}“brighten the night \\ for little bit”\end{tabular}                      & 0.33         & lights → night                                                        \\ \midrule
\begin{tabular}[c]{@{}c@{}}“please let me know \\ the alarm kept \\ for tuesday's meeting”\end{tabular} & \begin{tabular}[c]{@{}c@{}}“please let me know\\ the alarm cap \\ for tuesday's meet”\end{tabular} & 0.2          & \begin{tabular}[c]{@{}c@{}}kept → cap, \\ meeting → meet\end{tabular} \\ \bottomrule[1.2pt]
\end{tabular}
}
\caption{Some examples of clean and noisy ASR transcripts from the SLURP dataset. For ASR errors, the transcripts are tokenized, and the errors between the tokens are shown.}
\label{table_2}
\end{table}

In this section, we analyze the ASR error problems. We also introduce the noisy-channel model and the CCL method to address these issues. Our method performs the token-based contrastive learning followed by the consistency learning.

\subsection{Analyze ASR Errors}
\label{subsec:asrerror}
We analyze the ASR errors on the four benchmark datasets: SLURP, Timers, FSC, and SNIPS. We introduce the details of datasets in Appendix \ref{app:dataset}. The SLURP and FSC datasets have many classes, which can lead to ambiguities (such as \texttt{alarm\_set} and \texttt{alarm\_remove}) between them. We expect these datasets to be more sensitive to ASR errors than simpler ones.

Table \ref{table_1} shows the error rates according to various ASR modules. We use the noisy ASR transcripts provided by \citet{chang22c_interspeech} and \citet{kim2022improved} for SLURP and SNIPS, respectively. In addition, we extract the noisy ASR transcripts directly using Google API, HuBERT \cite{hsu2021hubert}, and Wav2vec 2.0 \cite{baevski2020wav2vec} speech recognizers on the remaining datasets. FSC has the lowest Word Error Rate (WER), whereas the SLURP and Timers datasets recorded by speakers of various accents have higher WER. Table \ref{table_2} lists some examples of the ASR errors such as (up, off), (light, night), etc. Unless these ASR errors are trained to the language model, a complete understanding of utterances is still limited. 

\subsection{Noisy-channel Model} 
\label{subsec:noisy}
We improve intent classification performance for the analyzed ASR errors based on the noisy-channel model. In the enhanced noisy-channel model, an intent probability given the noisy ASR transcript can be computed as a conditional probability for clean and noisy ASR transcripts as follows:
\begin{equation}
    \begin{split}
    P(I|\hat{X})&=\sum_{i}P(I, X_i|\hat{X}) \\
    &\approx P(I, X_c|\hat{X}) \\
    &=P(I|X_c,\hat{X})P(X_c|\hat{X}) \\
    \end{split}
\end{equation}
where among entire clean transcripts $X$, $X_c$ is the clean transcript corresponding to the noisy ASR transcript $\hat{X}$, and $I$ is the intent label. We compute an approximation by identifying the correlation between the noisy ASR transcript and the corresponding clean transcript rather than all clean transcripts. This shows that the intent probability for noisy ASR transcript can be maximized by the token alignment of two scripts and the similarity between two scripts. The reparameterized approximation identifies the corresponding clean transcript for the noisy ASR transcript and utilizes the associated clean transcript as a reference to comprehend the intent. 

\subsection{Token-based Contrastive Learning}
\label{subsec:tok}
We introduce the token-based contrastive learning that aligns clean and noisy ASR transcripts at the word and utterance token-levels. Token-based contrastive loss maximizes the probability $P(X_c|\hat{X})$ by aligning the corresponding word and entire utterance tokens. Given mini-batch $B$ of paired ($X_c$, $\hat{X}$), features are extracted by the respective encoders $E$ and $E^{\prime}$. Here, we use the pre-trained language model RoBERTa as the encoder. Inspired by \citep{oord2018representation, chen2020simple, gao2021simcse}, we define the contrastive loss as follows: 
\begin{equation}
\label{eqn:contrastive_loss}
\begin{array}{cl}
\normalsize\mathcal{\ell}(h, h^{\prime})=-\frac{1}{|\mathcal{P}|} \sum\limits_{i\in{\mathcal{P}}}\log{e^{(h_i^{\top}h_i^\prime)/\tau}\over{\sum\limits_{j\in\mathcal{N}}e^{(h_i^{\top}h_j^\prime)/\tau}}}
\end{array}
\end{equation}
where $h$ and $h^{\prime}$ are outputs of the encoders, $e$ is an exponential function, and $\tau$ is a hyperparameter. $\mathcal{P}$ is the set of the indices of the positive pairs, and $\mathcal{N}$ is the set of indices of all non-matching pairs.

\paragraph{Selective-token contrastive learning.} Unlike previous token-based contrastive learning \citep{su2022tacl}, since ASR errors are caused by insertion, deletion, and substitution, the noisy ASR transcript is not aligned with the clean transcript on the word token-level. We propose the selective-token contrastive learning to create pairs between word tokens via an edit distance algorithm. As shown in Figure \ref{figure2}, we select the positive pairs as the tokens with the minimum number of edits and the negative pairs as non-matching tokens in the batch. Given clean tokens $x_c$ (tokenized $X_c$) and noisy tokens $\hat{x}$ (tokenized $\hat{X}$), the proposed method is performed with respective outputs $z_c$ and $\hat{z}$ passing through the encoder. Selective-token contrastive loss that correlates the ASR errors for a token representation pair $(z_c, \hat{z})$ is defined as:
\begin{equation}
\label{eqn:sel-tokcont}
\normalsize\mathcal{L}_{sel}=\mathcal{\ell}(z_c,\hat{z}) + \mathcal{\ell}(z_c,z_c) 
\end{equation}
The representation $z_c$ for clean input contains semantic information for understanding the intent and thus should preserve the clean embeddings when pulling the positive pair close together. Therefore, by adding $\mathcal{\ell}(z_c,z_c)$, we train the reference network to maintain its expressiveness when receiving the clean transcript as input. Note that $\normalsize\mathcal{\ell}(z_c, z_c)$ is not equal to zero; this is attributable to the influence of the negative samples in the contrastive loss. According to \citet{wang2020understanding}, the contrastive loss minimizes the similarity between the features of the negative pair when positive pair features are perfectly aligned.

\paragraph{Utterance contrastive learning.} To align the error part of a given $(X_c, \hat{X})$ as the entire utterances, two transcripts are configured as positive pair and trained to increase similarity compared to negative pairs in the batch. Utterance contrastive loss is computed as follows using the [CLS] token representations $(Z_c, \hat{Z})$ that carry the implied meaning of the entire utterance:
\begin{equation}
\label{eqn:sel-tokcont}
\normalsize\mathcal{L}_{utt}=\mathcal{\ell}(Z_c,\hat{Z}) 
\end{equation} 
Finally, the token-based contrastive objective $\mathcal{L}_{ctr}$ is the weighted sum of the selective-token contrastive loss $\mathcal{L}_{sel}$ and the utterance contrastive loss $\mathcal{L}_{utt}$:
\begin{equation}
\label{eqn:sel-tokcont}
\normalsize\mathcal{L}_{ctr}=\lambda_{ctr}\mathcal{L}_{sel} + (1-\lambda_{ctr})\mathcal{L}_{utt} 
\end{equation}
where $\lambda_{ctr}$ is the hyperparameter to adjust the ratio of each loss.

\begin{table*}[!htp]
\setlength{\tabcolsep}{10pt}
\centering
\begin{tabular}{l|cc|cc|cc|cc}\toprule[1.2pt]
\multicolumn{1}{l|}{\multirow{2}{*}{method}} & \multicolumn{2}{c|}{Noisy$_{0.24}$} &
\multicolumn{2}{c|}{Noisy$_{0.44}$} & 
\multicolumn{2}{c|}{Noisy$_{0.56}$} & \multicolumn{2}{c}{Clean} \\
\multicolumn{1}{c|}{}             & Acc.    & F1    & Acc.    & F1   & Acc.    & F1  &Acc.    & F1 \\ \midrule
Clean-CE                           
& 82.6 & 82.8
& 64.7 & 60.51 
& 50.98 & 52.85        
& 96.35 & 96.25          \\ \midrule
Noisy-CE                           
& 85.19 & 85.23       
& 77.72 & 73.76  
& 70.53 & 70.65
& 96.3 & 96.29          \\ \midrule
SpokenCSE                        
& 85.26 & 85.66 
& 79.08 & \textbf{78.91}       
& 70.31 & 71.03 
& 95.82 & 96.21          \\ \midrule
\cellcolor[HTML]{EFEFEF}\textbf{CCL (Ours)} & \cellcolor[HTML]{EFEFEF}\textbf{86.22}    
& \cellcolor[HTML]{EFEFEF}\textbf{86.27} & \cellcolor[HTML]{EFEFEF}\textbf{81.51}  
& \cellcolor[HTML]{EFEFEF} 76.83 &\cellcolor[HTML]{EFEFEF}\textbf{73.12} & \cellcolor[HTML]{EFEFEF}\textbf{73.13}    
& \cellcolor[HTML]{EFEFEF}\textbf{96.99} & \cellcolor[HTML]{EFEFEF}\textbf{96.98} \\ \bottomrule[1.2pt]
\end{tabular}
\caption{Results for macro F1-score and accuracy performances on  SLURP dataset for our model trained with CCL method. We compare the CCL method and other baselines for clean and noisy transcripts.}
\label{table_3}
\end{table*}
\begin{table*}
\centering
\begin{tabular}{l|cccc|cccc}\toprule[1.2pt]
\multicolumn{1}{l|}{\multirow{2}{*}{method}} & \multicolumn{4}{c|}{Timers} & \multicolumn{4}{c}{FSC}  \\
\multicolumn{1}{c|}{}             
& Noisy$_{0.15}$  & Noisy$_{0.86}$  & Noisy$_{0.57}$     & Clean    
& Noisy$_{0.08}$ & Noisy$_{0.17}$    & Noisy$_{0.27}$    & Clean \\ \midrule
Clean-CE                           
& 99.69   & 94.87       & 89.15        & 100           
& 95.83   & 92.99     & 87.13       & \textbf{100}     \\ \midrule
Noisy-CE                           
& 99.97  & 99.66        & 99.29        & 100           
& 97.19   & 99.24     & 97.23       & 93.35          \\ \midrule
SpokenCSE                        
& 99.97       & 99.71   & 99.17        & 100           
& 99.18   & 99.24      & 98.5       & 98.02          \\ \midrule
\cellcolor[HTML]{EFEFEF}\textbf{CCL (Ours)} 
& \cellcolor[HTML]{EFEFEF} \textbf{99.98} & \cellcolor[HTML]{EFEFEF} \textbf{99.73} & \cellcolor[HTML]{EFEFEF} \textbf{99.45} & \cellcolor[HTML]{EFEFEF} \textbf{100}      
& \cellcolor[HTML]{EFEFEF} \textbf{99.31} & \cellcolor[HTML]{EFEFEF} \textbf{99.26} & \cellcolor[HTML]{EFEFEF} \textbf{98.87} & \cellcolor[HTML]{EFEFEF} 99.16 \\ \bottomrule[1.2pt] \end{tabular}
\caption{Intent classification accuracy results of Timers and FSC datasets.}
\label{table_4}
\end{table*}

\subsection{Consistency Learning}
\label{subsec:consistency}
We propose the consistency learning that maintains consistency for clean transcripts' latent features to understand better the intent of noisy ASR transcripts with severe recognition errors. In Figure \ref{figure1}, the consistency learning is used by adding a linear projection $f(.)$ and $f^{\prime}(.)$ to $E(.)$ and $E^{\prime}(.)$, respectively. Given the paired sentences ($X_c$, $\hat{X}$), the final outputs for the two samples are defined as:
\begin{equation}
    \begin{array}{cl}
        v_c&=f(E(X_c))\\
        \hat{v}&=f^{\prime}(E^{\prime}(\hat{X}))
    \end{array}
\end{equation}
where all tokens for the output of the encoder are flattened. The purpose of consistency learning is to prevent misclassification by additionally utilizing referenced $X_c$ when $\hat{X}$ is given as input. Increasing the consistency between referenced $X_c$ and $\hat{X}$ leads to the probability $P(I|X_c,\hat{X})$ being maximized. We reduce the distance between the noisy latent feature $\hat{v}$ and the referenced latent feature $v_c$ to mitigate the discrepancy. We are also consistent about the target probabilities of $p_c$ and $\hat{p}$ for $v_c$ and $\hat{v}$, respectively, to understand the target intent more precisely. By taking $\hat{X}$ as input and emphasizing the consistency of the latent features and target probabilities for $X_c$, we expect to avoid making incorrect predictions due to speech recognition errors. Formally, we derive the overall loss of consistency learning as:
\begin{equation}    
\normalsize\mathcal{L}_{con}=\lambda_{con}{(v_{c}-\hat{v})^2}+(1-\lambda_{con}){(\log{p_{c}}-\log{\hat{p}})^2} 
\end{equation}
where $\lambda_{con}$ is hyperparameter.

\paragraph{Reference network.} The reference latent feature used in the consistency learning should have a relatively better representation than the noisy latent feature. Therefore, the reference network is trained maximizing the probability $P(y|{X_c})$ of label $y$ over input $X_c$ at every step to be consistent with better latent features. After optimizing the reference network with cross-entropy loss at each step, the inference network bootstraps the optimized latent features.

\section{Experiment}
\subsection{Experiment Settings}
The proposed CCL is compared with the following three baseline methods:
\begin{itemize}
    \item SpokenCSE: To solve the ASR error problem of pre-trained RoBERTa, \citet {chang22c_interspeech} proposed contrastive learning between clean and noisy ASR transcripts. In the fine-tuning, self-distillation and supervised contrastive learning were applied to regulate the misprediction of noisy ASR transcripts. We conducted further experiments when results were unavailable for SpokenCSE.
    \item Clean-CE: The language model trained only on clean transcripts is vulnerable to noisy input. To investigate these vulnerabilities, we employ a technique that utilizes clean transcripts as inputs and fine-tunes them using cross-entropy (CE).
    \item Noisy-CE: Fine-tuning the noisy ASR transcript is the common method to solve the ASR error problem. While this method seems simple, it can be biased toward noisy ASR transcripts and lead to performance degradation of the clean transcript.
\end{itemize}
We conduct comparison experiments between baseline and CCL methods on SLURP, Timers, and FSC datasets. All models are based on RoBERTa to compare performance under the same conditions as \citet{chang22c_interspeech}. In addition, the CCL method is compared with various previous works in the field of SLU using SNIPS. We refer to the noisy ASR transcripts converted by different recognizers as Noisy$_{wer}$, and the clean transcripts as Clean. In the token-based contrastive learning, we use the same hyperparameters: learning rate 1e-6 and $\lambda_{ctr}$ 0.7. The learning rate is selected for consistency learning from \{1e-5, 3e-5, 5e-5, 7e-5, 1e-4\}. We tune $\lambda_{con}$ of best value that is 0.0 to 1.0. The batch size in our experiments is 64, the same for all. All experiments are performed with NVIDIA TITAN RTX 24GB GPU.

\subsection{Main Results}
Table \ref{table_3} lists the results of CCL and baseline methods on the SLURP test set. Clean-CE accurately predicts approximately half of the answers to Noisy$_{0.56}$, indicating that the language models pre-trained with only clean transcripts are susceptible to ASR errors. We also achieve the best results when measured by macro F1-score, which can consider the class imbalance of the SLURP dataset. Our method significantly outperforms Noisy-CE by 2.59\% (accuracy) and 2.48\% (F1-score) in Noisy$_{0.56}$. The performance gain is more significant for higher error rates than lower ones because our method aligns more ASR errors in noisy transcripts with clean ones containing contextual information to predict intent. Similar results are seen when running with four seeds to calculate the mean and standard deviation in Appendix Table \ref{seed}. Moreover, we attain the highest performance on clean transcripts, highlighting the robustness of the CCL method in handling speech recognition errors without compromising its performance on clean transcripts.  

The CCL and compared methods are evaluated in various settings. We employ two distinct datasets: simple Timers and complex FSC. Table \ref{table_4} lists the accuracy of clean and noisy ASR transcripts. Challenging tasks involving complex classes may exhibit heightened sensitivity to speech recognition errors, leading to performance degradation. However, we observe that the performance gain for complex FSC is more significant than that for simple Timers compared to the highest-performing method among the baselines. This indicates the proposed CCL method is robust to ASR errors under challenging tasks. 
\begin{figure*}[t]
    \centering
    \includegraphics[width=\textwidth]{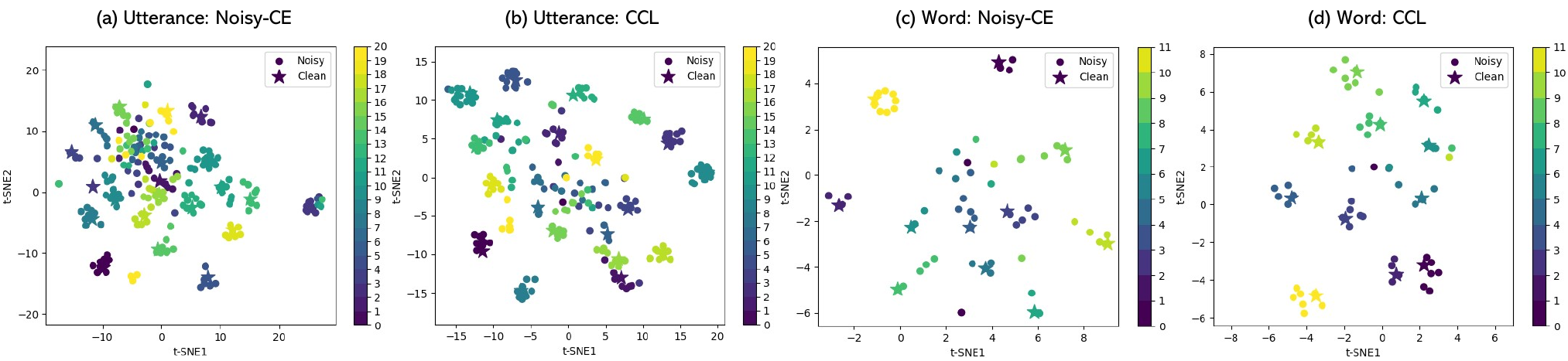}
    \caption{t-SNE visualization of clean and noisy ASR transcripts in word and utterance token-levels for SLURP test set. (a-b) each color represents a sample of clean transcript and associated noisy ASR transcripts. (c-d) each color indicates the tokens of sample, such as clean token and noisy ASR tokens.}
    \label{tsne}
\end{figure*}

Table \ref{table_5} presents the intent classification accuracy results for the SNIPS dataset. We compare the performance of various approaches in the SLU field. In end-to-end, \citet{lai2020towards} and \citet{kim2021two} employed training methods that simultaneously utilized audio and text data. Module-based approaches have attempted to address the ASR error problem using a method that increases the cosine similarity between two transcripts \citep{huang2020learning} or approaches that use phoneme sentences \citep{chen21g_interspeech}. Among the existing module-based approaches, the method with dual-BERT \citep{kim2022improved} achieved the best performance, although it was 0.03\% lower than the end-to-end approach. Whether end-to-end or module-based, our method achieves a higher performance of 98.72\%.
\begin{table}[t]
\centering
\setlength{\tabcolsep}{10pt}
\renewcommand{\arraystretch}{1.2}
\resizebox{\columnwidth}{!}{
\begin{tabular}{lcc}
\toprule[1.2pt]
approach                                                                      & Model & Acc.  \\
\hline
\multirow{2}{*}{End-to-end}                                                     & \citet {lai2020towards}   & 98.65 \\\cline{2-3}
                                                                            & \citet{kim2021two}   & 96.7  \\
\hline
\multirow{4}{*}{\begin{tabular}[c]{@{}c@{}}Module\end{tabular}} & \citet{huang2020learning}  & 89.55 \\\cline{2-3}
                                                                            & \citet{kim2022improved}   & 98.62 \\\cline{2-3}
                                                                            & \citet {chen21g_interspeech} & 85.4  \\\cline{2-3}
                                                                            & \textbf{CCL (Ours)}   & \textbf{98.72} \\
\bottomrule[1.2pt]
\end{tabular}}
\caption{Comparison of intent classification accuracy for SNIPS dataset between the end-to-end and the module-based approaches.}
\label{table_5}
\end{table}
\subsection{Further Analysis}
In this section, we analyze the factors that led the model trained using CCL to outperform the compared models. In addition, we conduct a range of experiments to verify the efficacy of the proposed CCL.
\begin{figure}[t]
    \centering
    \includegraphics[width=\columnwidth]{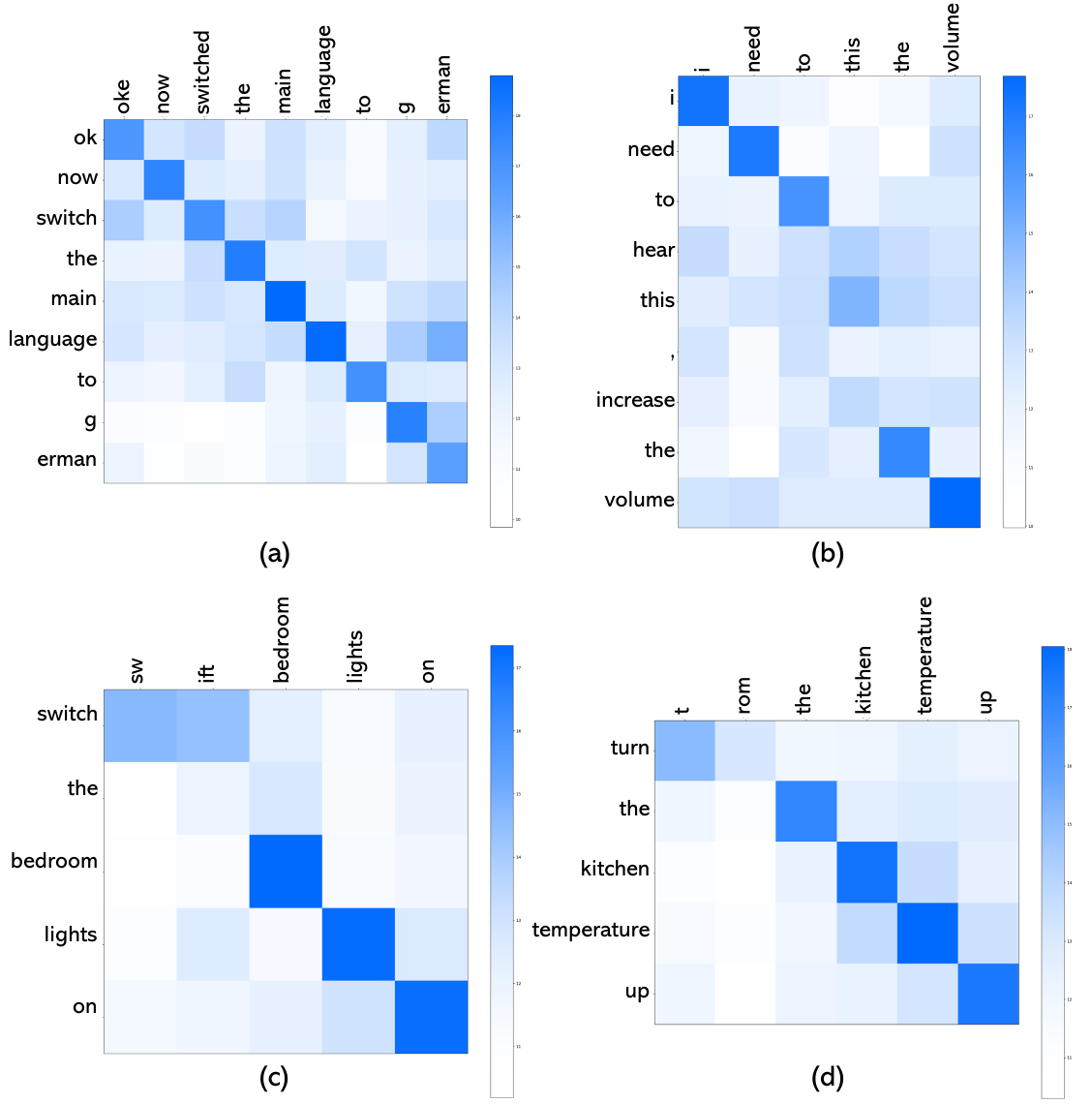}
    \caption{Token-based similarity map visualization results for FSC validation and test sets. X-axis and Y-axis represent clean and noisy ASR transcripts, respectively.}
    \label{Figure3}
    \label{figure_3}
\end{figure}
\paragraph{Token-based contrastive learning.} To better understand why CCL is robust to ASR errors, we visualize the outputs of encoders trained with CCL and Noisy-CE methods, respectively. We randomly select a sample (clean transcript, noisy ASR transcripts) in the SLURP test set with multiple speakers. Figure \ref{tsne} (a-b) visualizes the utterance token representations of multiple samples, indicating that the CCL method yields a much better separation than Noisy-CE; since our method can align several noisy ASR transcripts and clean transcript with high performance through the utterance contrastive learning. Figure \ref{tsne} (c-d) shows the results of sample containing pairs (clean word token, noisy word tokens). In CCL, the distance between clean and noisy tokens is reduced, demonstrating the effectiveness of selective-token contrastive learning. We analyze that the alignment of two transcripts in utterance and word token-levels leads to robust to ASR errors.

\begin{figure}[t]
    \centering
    \includegraphics[width=\columnwidth]{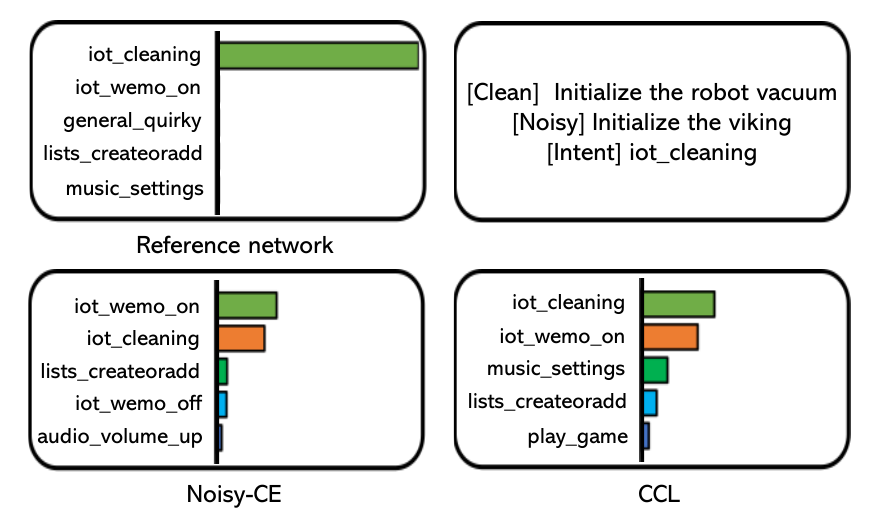}
    \caption{Top-5 intent prediction distributions on SLURP dataset from reference network, inference network (trained with CCL), and baseline model (trained Noisy-CE).}
    \label{fiture4}
\end{figure}

Figure \ref{Figure3} presents token-based similarity maps on the FSC valid and test sets. We observe that substituted tokens and aligned clean tokens have high similarity even when they have different meanings in Figure \ref{Figure3} (a). We can also see that clean and noisy ASR transcripts are semantically aligned when there are different token lengths caused by deletion (Figure \ref{Figure3} (b)). In contrast, Figure \ref{Figure3} (c-d) shows various speech recognition errors such as substitution, insertion, and deletion. The visualizations prove that selective-token contrastive learning works well in correlating the ASR errors, even when the tokens are not aligned or various ASR errors co-occur. Furthermore, we describe the analysis of the utterance-based similarity maps in  Appendix \ref{sec:app_utt}.
\paragraph{Consistency learning.} We explore the advantages of the proposed consistency learning over the cross-entropy loss commonly used in fine-tuning. Figure \ref{fiture4} shows the probability distribution of the SLURP test set for the reference network, inference network (trained with CCL), and baseline model (trained Noisy-CE). The probability distribution of the reference network is the result for the clean transcript as input, and the rest are the results for the noisy ASR transcript. The top-5 probability values out of a set of 60 classes are depicted in all the visualizations. While the reference network accurately predicts the answer \texttt{iot\_cleaning}, Noisy-CE encounters confusion between \texttt{iot\_cleaning} and \texttt{iot\_wemo\_on} for the noisy ASR transcript where important keywords were deleted and replaced. On the other hand, the CCL method exhibits a similar level of confusion as Noisy-CE, but this method allows the model to predict the correct answer. We analyze that the CCL method mitigates misclassifications by maintaining consistency with the latent feature of the reference network rather than relying on a fixed target. 

\paragraph{Ablation studies.} We conduct an ablation study to investigate the impact of the losses used in the CCL method. Table \ref{table_ab} demonstrates that the CCL method achieves the best accuracy. We observe that the higher the speech recognition error rate, the more significant the performance degradation while removing the token-based contrastive loss. This result shows that the proposed token-based contrastive learning is increasingly effective in noisy environments. In token-based contrastive learning, better performance is observed when mismatched tokens are selectively correlated using $\normalsize\mathcal{L}_{sel}$, rather than $\normalsize\mathcal{L}_{utt}$. In the fine-tuning, we attain improved results employing the consistency loss rather than the commonly used cross-entropy loss. This highlights the significance of preserving consistency between clean and noisy latent representations. We conduct additional ablation experiments using cross-entropy loss instead of consistency objective in Appendix \ref{sec:app_ablation}.

\begin{table}[t!]
\renewcommand{\arraystretch}{1.5}
\resizebox{\columnwidth}{!}{
\begin{tabular}{c|cc|cc}
\toprule[1.2pt]
\multicolumn{1}{c|}{\multirow{2}{*}{method}}               & \multicolumn{2}{c|}{SLURP}                                                                                    & \multicolumn{2}{c}{Timers}                                                                           \\
\multicolumn{1}{c|}{}                                & \begin{tabular}[c]{@{}c@{}}Noisy$_{0.24}$ \end{tabular} & \begin{tabular}[c]{@{}c@{}}Noisy$_{0.56}$ \end{tabular} & \begin{tabular}[c]{@{}c@{}}Noisy$_{0.15}$ \end{tabular} & \begin{tabular}[c]{@{}c@{}}Noisy$_{0.57}$ \end{tabular} \\
\midrule
$\mathcal{L}_{ce}$                                                    & 85.19                                                 & 70.53                                                 & 99.97                                                & 99.29                                                 \\
\midrule
$\mathcal{L}_{con}$                                                    & 85.95                                                & 72.81                                                 & 99.88                                                & 99.26                                                 \\
\midrule
\begin{tabular}[c]{@{}c@{}} $\mathcal{L}_{con}$ \\ + $\mathcal{L}_{sel}$\end{tabular}     & 86.17                                                & 72.3                                                  & 99.89                                                & 99.26                                                 \\
\midrule
\begin{tabular}[c]{@{}c@{}}$\mathcal{L}_{con}$ \\ + $\mathcal{L}_{utt}$\end{tabular}      & 86.13                                                & 72.16                                                 & 99.89                                                & 99.24                                                 \\
\midrule
\begin{tabular}[c]{@{}c@{}}$\mathcal{L}_{con}$ \\ + $\mathcal{L}_{sel}$\\ + $\mathcal{L}_{utt}$ \end{tabular} & \textbf{86.22}                                                & \textbf{73.12}                                                 & \textbf{99.98}                                                & \textbf{99.45}  \\
\bottomrule[1.2pt]
\end{tabular}
}
\caption{Ablation study of different losses.}
\label{table_ab}
\end{table}
\begin{table}[t]
\renewcommand{\arraystretch}{1.5}
\centering
\resizebox{\columnwidth}{!}{
\begin{tabular}{@{}c|c|c|c@{}}
\toprule[1.2pt]
\textbf{Noisy ASR transcript} & \textbf{GPT4} & \textbf{CCL} & \textbf{GT} \\ \midrule
dilliitalan                    & iot\_hue\_lightdim & alarm\_remove & alarm\_remove \\
the lid alam                   & alarm\_set         & alarm\_remove & alarm\_remove \\
donald sultan book up          & play\_audiobook    & qa\_factoid   & qa\_factoid   \\ \bottomrule[1.2pt]
\end{tabular}
}
\caption{Comparative analysis of GPT4 and CCL} 
\label{gpt4}
\end{table}

\subsection{Comparision with Large Language Models}

\begin{figure}[t!]
    \centering
    \includegraphics[width=\columnwidth]{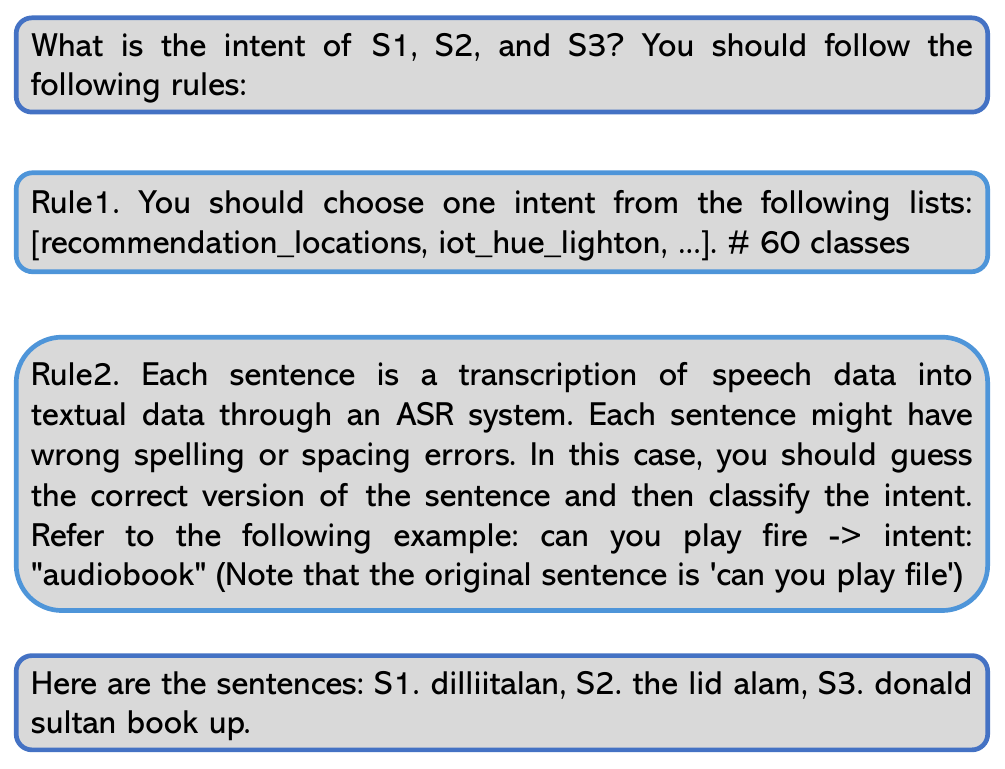}
    \caption{Prompt structure for LLMs.}
    \label{ap_prompt}
\end{figure}

Recent advances in Large Language Models (LLMs) have attracted significant research interest due to their representation power, which has led to notable successes in natural language processing \cite{ouyang2022training, achiam2023gpt, touvron2023llama}, multi-modal \cite{li2023blip, lin2023video}, and computer vision \cite{zhao2023antgpt, wang2024visionllm} fields. However, it remains unclear whether LLMs trained only on the clean text can identify and contextually understand the ASR errors. We conduct a comparison experiment on GPT4 \cite{achiam2023gpt} and our CCL method to verify the ability to understand the errors. Through this experiment, we aim to observe how each method classifies intent labels on the same noisy ASR transcripts from the SLURP dataset. For GPT4, we provided the prompt as shown in Figure \ref{ap_prompt}. As shown in the Table \ref{gpt4}, GPT-4 can understand the intent \texttt{alarm} in the transcript containing the minor ASR errors. However, it still has trouble detecting severe ASR errors, which could lead to misclassification. In contrast, CCL detects a wide range of ASR errors, even if they are hard to understand, allowing our method to predict precise intention.

As mentioned above, LLMs demonstrate constraints in understanding the ASR errors without additional training. We fine-tuned the decoder-based and encoder-based LLMs using either the cross-entropy loss or the CCL method. Due to the resource limitation, we use the GPT-Neo \cite{gpt-neo} and GPT-2 \cite{radford2019language} models. Table \ref{tab:dec} shows the results of decoder-based and encoder-based models on the SLURP test set. The encoder-based language model consistently outperforms decoder-based language models, even with fewer parameters (110M). We believe the encoder structure is apt for classification tasks because the model can capture contextual information in bi-directions. Furthermore, our CCL method enhances the performance of the encoder- and decoder-based language models in extremely noisy environments.

\section{Conclusion}
We have introduced CCL, a novel method to solve the ASR error problem of pre-trained language models. We designed the two-stage method: (i) token-based contrastive learning to correlate the ASR errors at both the word and utterance token-levels and (ii) consistency learning to maintain coherence with the latent feature of relatively better clean transcripts. Experiments and analysis on four benchmark datasets have demonstrated that our model significantly outperforms previous models in various speech recognition error rates.  

\section{Limitations}
\begin{table}[t!]
\centering
\renewcommand{\arraystretch}{1.2}
\resizebox{0.8\columnwidth}{!}{
\begin{tabular}{l|c|c}
\toprule[1.2pt]
model & Noisy$_{0.24}$ & Noisy$_{0.56}$ \\ \midrule
GPT-Neo (125M) & 76.55 & 57.44 \\
GPT-Neo+CCL & 75.99 & 64.48 \\
GPT-2 (355M) & 83.17 & 65.94 \\
GPT-2+CCL & 80.33 & 70.23 \\
RoBERTa (110M) & 85.19 & 70.53 \\
RoBERTa+CCL & \textbf{86.22} & \textbf{73.12} \\
\bottomrule[1.2pt]
\end{tabular}}
\caption{Performance of various models on the SLURP benchmark dataset.}
\label{tab:dec}
\end{table}
Although our work makes further progress in the ASR problem of SLU, it is subject to two potential limitations. First, when ASR errors seriously occur that the original meaning is unrecognizable, the model trained with CCL predicts wrong answers as in Appendix \ref{Qualitative results}. In the future work, we plan to develop a method to address the mismatch between noisy ASR transcripts and labels. Then, there has been a growing interest in the field of multilingual, but most ASR error studies have centered on English. We need to expand the ASR error problem to multilingual in the future work. 

\section{Acknowledgements}
This research was supported by the MSIT(Ministry of Science and ICT), Korea, under the ITRC(Information Technology Research Center) support program(IITP-2024-2020-0-01808) supervised by the IITP(Institute of Information \& Communications Technology Planning \& Evaluation).

\appendix
\clearpage
\section{Dataset Description}
\label{app:dataset}
We experiment with the SLU benchmark datasets: 
\begin{itemize}
    \item SLURP: The SLURP dataset, a challenging SLU benchmark dataset, consists of house robot voice commands from 177 speakers. Since SLURP contains inaccurate labels, we use a dataset from \citet{chang22c_interspeech} where the labels have been denoised. We use 60 intent labels and divide each dataset into 50,627, 13,078, and 10,992 for train, valid, and test, respectively. 
    \item Timers: The Timers dataset is simple data recorded from 95 speakers, smaller than SLURP. This dataset consists of four labels: SetAlarm, SetTimer, SimpleMath, and UnitConversion. We split the data into training, validation, and testing of 7,000, 1,000, and 12,000, respectively.
    \item FSC: The FSC dataset consists of commands for smart-home or voice assistants of 97 speakers. As in \citet{lugosch2019speech}, we use 31 intents, 23,132 training, 3,118 validation, and 3,793 testing.
    \item SNIPS: A well-known text-based intent classification SNIPS dataset consists of 13,084 training, 700 validation, and 700 evaluation. We classify seven intents using an ASR transcript from \citet{kim2022improved}, which transcribed synthetic speech files into text. 
\end{itemize}

\section{Additional Analysis}
\label{app:Additional Analysis}
\begin{figure}[t!]
    \centering
    \includegraphics[width=\columnwidth]{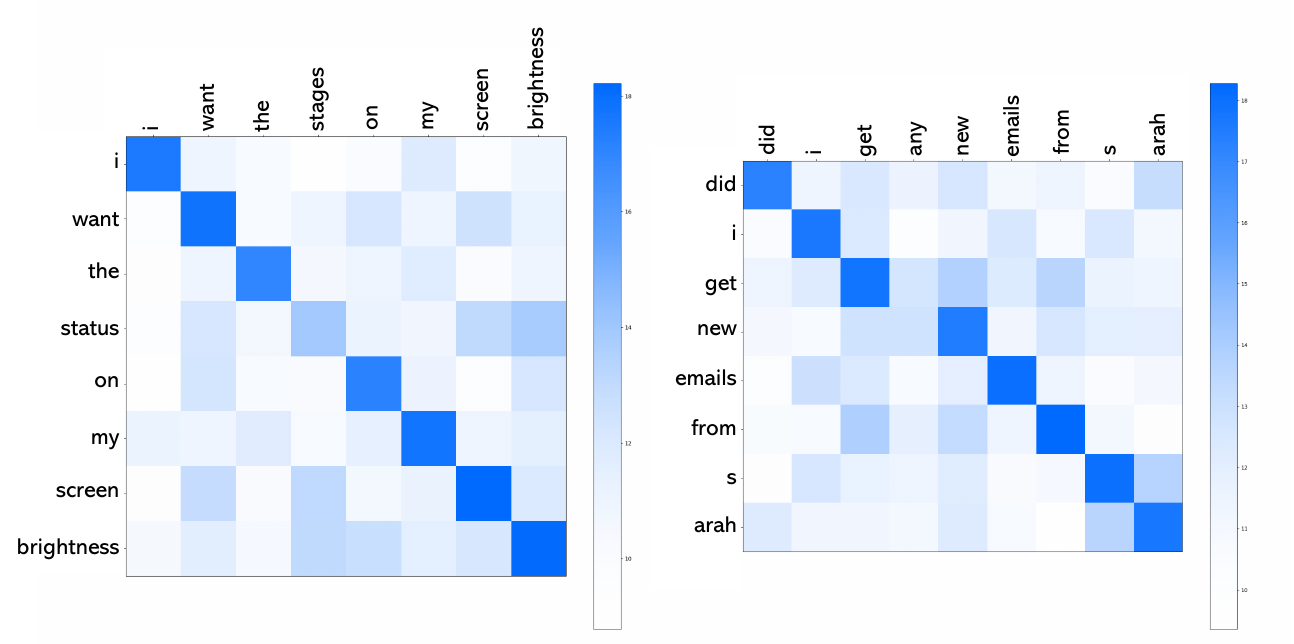}
    \caption{Visualization results of further similarity maps between tokens.}
    \label{ap_tok}
\end{figure}
\subsection{Selective-token Contrastive Learning}
\label{sec:app_sel}
Figure \ref{ap_tok} shows the further examples of similarity maps that visually represents the similarity between the outputs of the inference and reference networks trained using the token-based contrastive loss. The visualization experiment uses the SLURP test set. In Figure \ref{ap_tok} (left), a recognition error occurs for clean transcript “\texttt{I want the status on my screen brightness}” and noisy ASR transcript “\texttt{I want the stages on my screen brightness}”, where \texttt{status} is replaced by \texttt{stages}. Despite \texttt{status} and \texttt{stages} being distinct tokens, their similarity scores toward each other are high. In Figure \ref{ap_tok} (right), we can observe that the insertion error token \texttt{any} is not aligned with the other tokens. These visualizations demonstrate that we can identify the recognition errors even while handling unfamiliar data by employing the selective-token contrastive learning. 

\begin{table}[t!]
\setlength{\tabcolsep}{10pt}
\renewcommand{\arraystretch}{1.5}
\resizebox{\columnwidth}{!}{
\begin{tabular}{l|cc|cc}\toprule[1.2pt]
\multicolumn{1}{l|}{\multirow{2}{*}{method}}               & \multicolumn{2}{c|}{SLURP}                                                                                    & \multicolumn{2}{c}{Timers}                                                                           \\
\multicolumn{1}{l|}{}                                & \begin{tabular}[c]{@{}c@{}}Noisy$_{0.24}$ \end{tabular} & \begin{tabular}[c]{@{}c@{}}Noisy$_{0.56}$ \end{tabular} & \begin{tabular}[c]{@{}c@{}}Noisy$_{0.15}$ \end{tabular} & \begin{tabular}[c]{@{}c@{}}Noisy$_{0.57}$ \end{tabular} \\
\midrule
$\mathcal{L}_{ce}$                                                    & 85.19                                                 & 70.53                                                 & 99.97                                                & 99.29                                                 \\
\midrule

\begin{tabular}[c]{@{}c@{}} $\mathcal{L}_{ce}$ \\ + $\mathcal{L}_{sel}$\end{tabular}     & 84.61                                                & 70.52                                                  & 99.98                                                & \textbf{99.73}                                                 \\
\midrule

\begin{tabular}[c]{@{}c@{}}$\mathcal{L}_{ce}$ \\ + $\mathcal{L}_{utt}$\end{tabular}      & 84.66                                                & 70.61                                                & 99.97                                                & 99.66                                                 \\
\midrule
\begin{tabular}[c]{@{}c@{}}$\mathcal{L}_{ce}$ \\ + $\mathcal{L}_{sel}$\\ + $\mathcal{L}_{utt}$ \end{tabular} & 84.92                                                & 71.02                                               & 99.98                                                & 99.67  \\
\midrule
$\mathcal{L}_{cons}$                                                    & 85.95                                                & 72.81                                                 & 99.88                                                & 99.26                                                 \\
\midrule
\begin{tabular}[c]{@{}c@{}} $\mathcal{L}_{cons}$ \\ + $\mathcal{L}_{sel}$\end{tabular}     & 86.17                                                & 72.3                                                  & 99.89                                                & 99.26                                                 \\
\midrule
\begin{tabular}[c]{@{}c@{}}$\mathcal{L}_{cons}$ \\ + $\mathcal{L}_{utt}$\end{tabular}      & 86.13                                                & 72.16                                                 & 99.89                                                & 99.24                                                 \\
\midrule
\begin{tabular}[c]{@{}c@{}}$\mathcal{L}_{cons}$ \\ + $\mathcal{L}_{sel}$\\ + $\mathcal{L}_{utt}$ \end{tabular} & \textbf{86.22}                                                & \textbf{73.12}                                                 & \textbf{99.98}                                                & 99.45  \\
\bottomrule[1.2pt]
\end{tabular}
}
\caption{Total ablation studies of different losses.}
\label{table_7}
\end{table}
\begin{figure*}[t]
    \centering
    \includegraphics[width=0.8\textwidth]{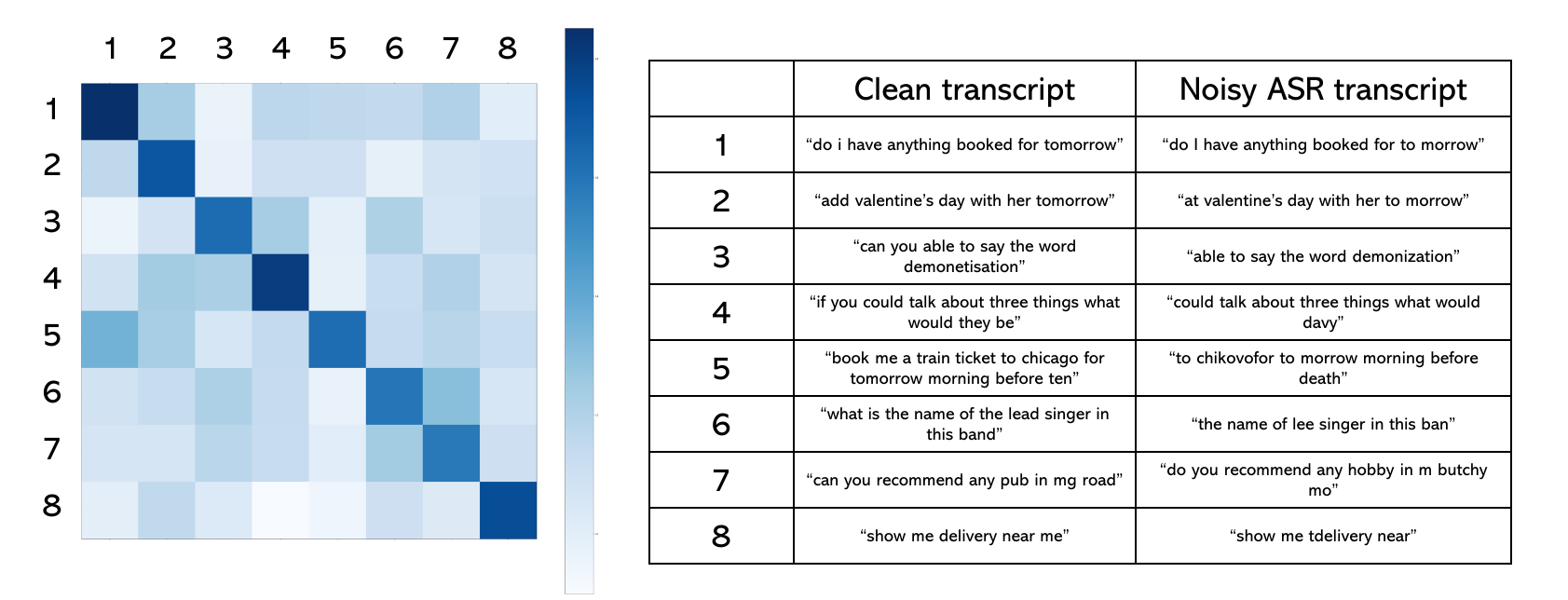}
    \caption{Left side shows the similarity map visualization between randomly sampled examples from the SLURP test set. Details of the sampled example are on the right side.}
    \label{ap_utt}
\end{figure*}

\subsection{Utterance Contrastive Learning}
\label{sec:app_utt}
We compare the contextual representations of clean and noisy ASR transcripts to demonstrate that the utterance contrastive objective works well. To quantify contextual similarity, we calculate the similarity score between [CLS] tokens in each of the reference and inference networks. Figure \ref{ap_utt} shows the similarity map for samples, where the x-axis and y-axis represent noisy ASR and clean transcripts, respectively. Overall, the diagonal similarity scores are higher than the other parts. In particular, the clean transcript “\texttt{Book me a train ticket to chicago for tomorrow morning before ten}” and the noisy ASR transcript “\texttt{To chikovofor to tomorrow morning before death}” have high similarity scores, even though the speech recognition error has lost essential keywords. This indicates that contrastive learning at the utterance level is effective in implicitly detecting contextual disparities between two transcripts.
\begin{table}[t!]
\renewcommand{\arraystretch}{1.2}
\resizebox{\columnwidth}{!}{
\begin{tabular}{c|c|c|c}
\toprule[1.2pt]
\textbf{Clean transcript} & \textbf{Noisy ASR transcript} & \textbf{PR}              & \textbf{GT}                 \\ \midrule
health                    & naws                          & news\_query     & general\_quirky    \\
                      
please turn lights off    & you send the baxle            & transport\_taxi & iot\_hue\_lightoff \\

create a new list         & read in yolist                & lists\_query    & lists\_createoradd \\ \midrule

delete alarm              & dilliitalan                   & alarm\_remove   & alarm\_remove      \\

delete alarm              & the lid alam                  & alarm\_remove   & alarm\_remove      \\

what time is it           & what's im he say              & datetime\_query & datetime\_query    \\ \bottomrule[1.2pt]
\end{tabular}}
\caption{Intent classification results of the CCL method for noisy ASR transcripts on the SLURP dataset.}
\label{t2}
\end{table}
\subsection{Ablation studies}
\label{sec:app_ablation}
Table \ref{table_7} shows the results of additional ablation experiments with cross-entropy loss while retaining token-based contrastive learning. In most cases, token-based contrastive learning typically demonstrates enhanced performance by employing the consistency loss instead of cross-entropy loss. This highlights the significance of preserving the consistency between clean and noisy latent representations, particularly in environments where speech recognition errors are prevalent. On the other hand, the best performance is achieved on Noisy$_{0.57}$ when the cross-entropy loss and the selective-contrastive loss are used together. We analyze that cross-entropy loss works well in the simple task with four intents. However, we need the consistency learning that works well on complex tasks to capture the diverse intents of users in the real world.

\subsection{Qualitative results}
\label{Qualitative results}
Table \ref{t2} lists the intent prediction results on the SLURP test set. From rows 1-3, the proposed CCL predicts intent that is different from the ground truth; however, this is matched when considered in the context of a noisy ASR transcript. In other words, our method tends to predict incorrect answers when the original meaning is lost in the speech recognition process. Although such results with 1.0 WER are rare in the SLU benchmark dataset, we need to improve these issues for real-world use in noisy environments. For future work, we will develop a method that addresses the mismatch between noisy ASR transcripts and their corresponding labels. As shown in rows 4-6, our method accurately predicts intent class when the original meaning is partially preserved.

\begin{table}[t!]
\centering
\setlength{\tabcolsep}{10pt}
\renewcommand{\arraystretch}{1.2}
\resizebox{0.7\columnwidth}{!}{
\begin{tabular}{l|c|c}
\toprule[1.2pt]
method & Noisy$_{0.24}$ & Noisy$_{0.56}$ \\ \midrule
CN-CE & 84.86 & 71.12 \\
SimCSE & 83.51 & 70.71 \\
CCL (Ours) & \textbf{86.22} & \textbf{73.12} \\
\bottomrule[1.2pt]
\end{tabular}}
\caption{Performance of further baselines on SLURP benchmark.}
\label{furbaseline}
\end{table}
\begin{table*}[h!]
\centering
\begin{adjustbox}{max width=\textwidth}
\begin{tabular}{l|ccc|ccc|ccc} 
\toprule[1.2pt]
& \multicolumn{3}{c|}{SLURP} & \multicolumn{3}{c|}{Timers} & \multicolumn{3}{c}{FSC} \\ 
\multirow{-2}{*}{method} & Noisy$_{0.24}$       & Noisy$_{0.44}$       & Noisy$_{0.56}$      & Noisy$_{0.15}$       & Noisy$_{0.86}$       & Noisy$_{0.57}$       & Noisy$_{0.08}$      & Noisy$_{0.17}$      & Noisy$_{0.27}$      \\ \midrule
Clean-CE                                          
& 83.43$_{\pm0.55}$      & 64.67$_{\pm0.03}$     & 52.13$_{\pm0.79}$    
& 99.70$_{\pm0.07}$      & 93.37$_{\pm4.05}$      & 90.09$_{\pm3.05}$      
& 93.84$_{\pm1.33}$     & 91.36$_{\pm1.82}$     & 86.87$_{\pm1.18}$     \\ \midrule
Noisy-CE                                          
& 84.97$_{\pm0.20}$      & 79.04$_{\pm0.88}$      & 70.44$_{\pm0.08}$     
& \textbf{99.98$_{\pm0.01}$}      & 99.69$_{\pm0.02}$      & 99.09$_{\pm0.15}$      
& 96.01$_{\pm0.92}$     & 99.09$_{\pm0.14}$     & 98.28$_{\pm0.70}$     \\ \midrule
SpokenCSE                                         
& 85.21$_{\pm0.24}$      & 78.72$_{\pm0.43}$      & 69.72$_{\pm0.40}$     
& 99.97$_{\pm0.01}$      & \textbf{99.73$_{\pm0.02}$}      & 99.24$_{\pm0.08}$      
& 96.16$_{\pm3.44}$     & 99.19$_{\pm0.07}$     & 98.53$_{\pm0.05}$    \\ \midrule
CCL (Ours)                                        
& \textbf{86.28$_{\pm0.10}$}  & \textbf{81.45$_{\pm0.13}$}      & \textbf{72.69$_{\pm0.30}$}   
& 99.97$_{\pm0.01}$      & 99.64$_{\pm0.12}$      & \textbf{99.33$_{\pm0.32}$}      
& \textbf{99.23$_{\pm0.06}$ }    & \textbf{99.32$_{\pm0.05}$}     & \textbf{98.77$_{\pm0.11}$}     \\ \bottomrule[1.2pt]
\end{tabular}
\end{adjustbox}
\caption{Performance comparison results with the average and standard deviation on the noisy ASR transcripts from SLURP test set. Each result is calculated across 4 random seeds. We denote optimal performance in bold.}
\label{seed}
\end{table*}

\begin{figure*}[h!]
    \centering
    \includegraphics[width=\textwidth]{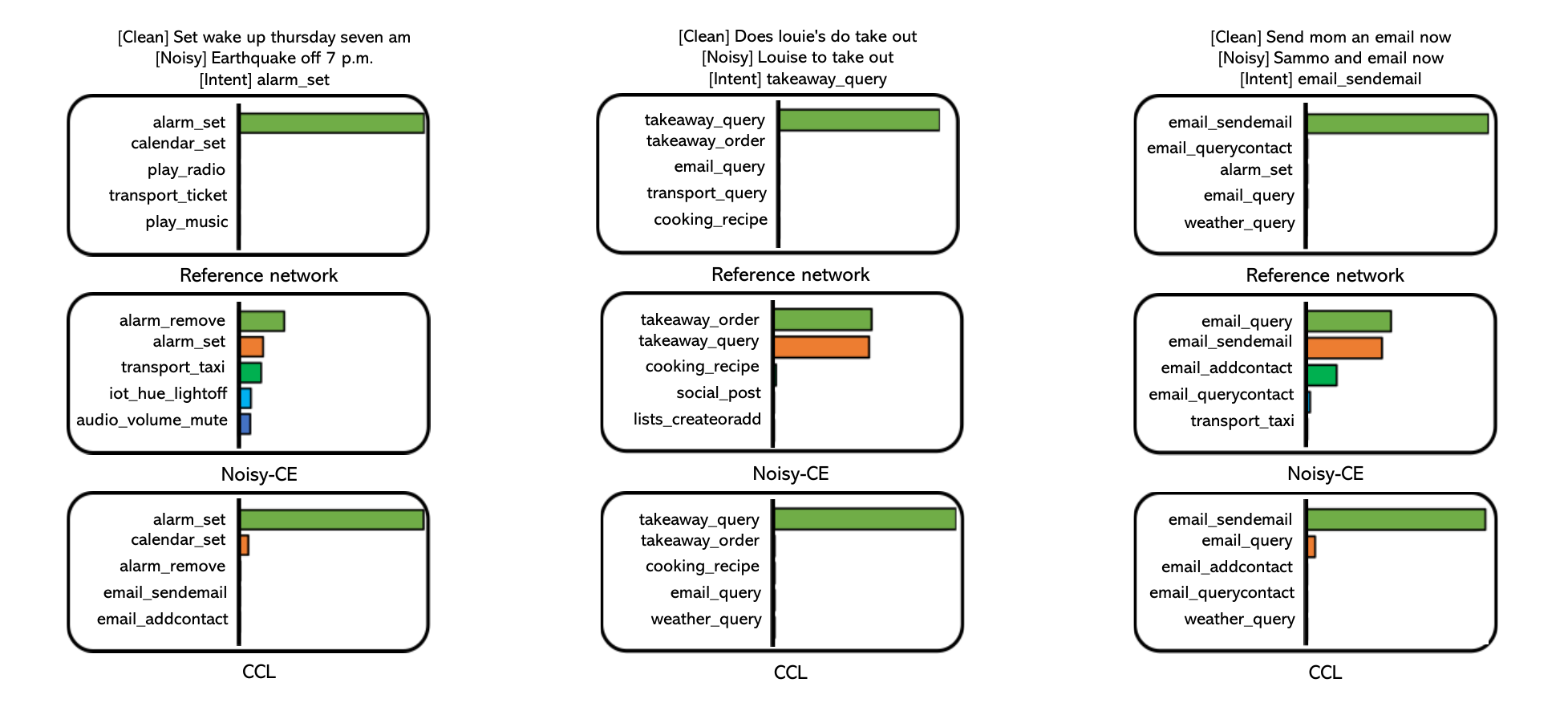}
    \caption{Comparison of prediction distributions on the SLURP test set from reference networks, inference networks, and Noisy-Net.}
    \label{ap_consistency}
\end{figure*}
\section{Comparison with further baselines}
We compare the CCL method with the following baselines:
\begin{itemize}
    \item CN-CE: A simple baseline is to train a model on the combination of clean and noisy ASR transcripts. This might make it more reasonable than another baseline since it utilizes both clean and noisy data, which is consistent with the CCL method.
    \item SimCSE \cite{gao2021simcse}: By applying contrastive learning, the method significantly advances the state-of-the-art in sentence embedding. The method maximizes the similarity between the original and randomly dropped features. We employ the identical noisy ASR transcripts to serve as a positive pair.
\end{itemize}
Table \ref{furbaseline} shows that the CCL and CN-CE methods using both clean and noisy ASR transcripts achieve better results than SimCSE only trained with noisy ASR transcripts. Furthermore, our method, which focuses on correlating errors and emphasizing consistency between two transcripts, is better suited to noisy environments than the simple CN-CE method that employs the cross-entropy loss to train on a fixed target.


\begin{thebibliography}{40}
\expandafter\ifx\csname natexlab\endcsname\relax\def\natexlab#1{#1}\fi

\bibitem[{Achiam et~al.(2023)Achiam, Adler, Agarwal, Ahmad, Akkaya, Aleman, Almeida, Altenschmidt, Altman, Anadkat et~al.}]{achiam2023gpt}
Josh Achiam, Steven Adler, Sandhini Agarwal, Lama Ahmad, Ilge Akkaya, Florencia~Leoni Aleman, Diogo Almeida, Janko Altenschmidt, Sam Altman, Shyamal Anadkat, et~al. 2023.
\newblock Gpt-4 technical report.
\newblock \emph{arXiv preprint arXiv:2303.08774}.

\bibitem[{Baevski et~al.(2020)Baevski, Zhou, Mohamed, and Auli}]{baevski2020wav2vec}
Alexei Baevski, Yuhao Zhou, Abdelrahman Mohamed, and Michael Auli. 2020.
\newblock wav2vec 2.0: A framework for self-supervised learning of speech representations.
\newblock \emph{Advances in neural information processing systems}, 33:12449--12460.

\bibitem[{Bahdanau et~al.(2015)Bahdanau, Cho, and Bengio}]{bahdanau2015neural}
Dzmitry Bahdanau, Kyung~Hyun Cho, and Yoshua Bengio. 2015.
\newblock Neural machine translation by jointly learning to align and translate.
\newblock In \emph{3rd International Conference on Learning Representations, ICLR 2015}.

\bibitem[{Bastianelli et~al.(2020)Bastianelli, Vanzo, Swietojanski, and Rieser}]{bastianelli-etal-2020-slurp}
Emanuele Bastianelli, Andrea Vanzo, Pawel Swietojanski, and Verena Rieser. 2020.
\newblock \href {https://doi.org/10.18653/v1/2020.emnlp-main.588} {{SLURP}: A spoken language understanding resource package}.
\newblock In \emph{Proceedings of the 2020 Conference on Empirical Methods in Natural Language Processing (EMNLP)}, pages 7252--7262, Online. Association for Computational Linguistics.

\bibitem[{Black et~al.(2021)Black, Gao, Wang, Leahy, and Biderman}]{gpt-neo}
Sid Black, Leo Gao, Phil Wang, Connor Leahy, and Stella Biderman. 2021.
\newblock \href {https://doi.org/10.5281/zenodo.5297715} {{GPT-Neo: Large Scale Autoregressive Language Modeling with Mesh-Tensorflow}}.
\newblock {If you use this software, please cite it using these metadata.}

\bibitem[{Cao et~al.(2021)Cao, Potdar, and Avila}]{cao2021sequential}
Yiran Cao, Nihal Potdar, and Anderson~R Avila. 2021.
\newblock Sequential end-to-end intent and slot label classification and localization.
\newblock \emph{arXiv preprint arXiv:2106.04660}.

\bibitem[{Chang and Chen(2022)}]{chang22c_interspeech}
Ya-Hsin Chang and Yun-Nung Chen. 2022.
\newblock \href {https://doi.org/10.21437/Interspeech.2022-781} {{Contrastive Learning for Improving ASR Robustness in Spoken Language Understanding}}.
\newblock In \emph{Proc. Interspeech 2022}, pages 3458--3462.

\bibitem[{Chen et~al.(2021)Chen, Wang, and Zhang}]{chen21g_interspeech}
Qian Chen, Wen Wang, and Qinglin Zhang. 2021.
\newblock \href {https://doi.org/10.21437/Interspeech.2021-234} {{Pre-Training for Spoken Language Understanding with Joint Textual and Phonetic Representation Learning}}.
\newblock In \emph{Proc. Interspeech 2021}, pages 1244--1248.

\bibitem[{Chen et~al.(2019)Chen, Zhuo, and Wang}]{chen2019bert}
Qian Chen, Zhu Zhuo, and Wen Wang. 2019.
\newblock Bert for joint intent classification and slot filling.
\newblock \emph{arXiv preprint arXiv:1902.10909}.

\bibitem[{Chen et~al.(2020)Chen, Kornblith, Norouzi, and Hinton}]{chen2020simple}
Ting Chen, Simon Kornblith, Mohammad Norouzi, and Geoffrey Hinton. 2020.
\newblock A simple framework for contrastive learning of visual representations.
\newblock In \emph{International conference on machine learning}, pages 1597--1607. PMLR.

\bibitem[{Chen et~al.(2018)Chen, Price, and Bangalore}]{chen2018spoken}
Yuan-Ping Chen, Ryan Price, and Srinivas Bangalore. 2018.
\newblock Spoken language understanding without speech recognition.
\newblock In \emph{2018 IEEE International Conference on Acoustics, Speech and Signal Processing (ICASSP)}, pages 6189--6193. IEEE.

\bibitem[{Coucke et~al.(2018)Coucke, Saade, Ball, Bluche, Caulier, Leroy, Doumouro, Gisselbrecht, Caltagirone, Lavril et~al.}]{coucke2018snips}
Alice Coucke, Alaa Saade, Adrien Ball, Th{\'e}odore Bluche, Alexandre Caulier, David Leroy, Cl{\'e}ment Doumouro, Thibault Gisselbrecht, Francesco Caltagirone, Thibaut Lavril, et~al. 2018.
\newblock Snips voice platform: an embedded spoken language understanding system for private-by-design voice interfaces.
\newblock \emph{arXiv preprint arXiv:1805.10190}.

\bibitem[{Devlin et~al.(2018)Devlin, Chang, Lee, and Toutanova}]{devlin2018bert}
Jacob Devlin, Ming-Wei Chang, Kenton Lee, and Kristina Toutanova. 2018.
\newblock Bert: Pre-training of deep bidirectional transformers for language understanding.
\newblock \emph{arXiv preprint arXiv:1810.04805}.

\bibitem[{Gao et~al.(2021)Gao, Yao, and Chen}]{gao2021simcse}
Tianyu Gao, Xingcheng Yao, and Danqi Chen. 2021.
\newblock Simcse: Simple contrastive learning of sentence embeddings.
\newblock In \emph{Proceedings of the 2021 Conference on Empirical Methods in Natural Language Processing}, pages 6894--6910.

\bibitem[{Goo et~al.(2018)Goo, Gao, Hsu, Huo, Chen, Hsu, and Chen}]{goo2018slot}
Chih-Wen Goo, Guang Gao, Yun-Kai Hsu, Chih-Li Huo, Tsung-Chieh Chen, Keng-Wei Hsu, and Yun-Nung Chen. 2018.
\newblock Slot-gated modeling for joint slot filling and intent prediction.
\newblock In \emph{Proceedings of the 2018 Conference of the North American Chapter of the Association for Computational Linguistics: Human Language Technologies, Volume 2 (Short Papers)}, pages 753--757.

\bibitem[{Hsu et~al.(2021)Hsu, Bolte, Tsai, Lakhotia, Salakhutdinov, and Mohamed}]{hsu2021hubert}
Wei-Ning Hsu, Benjamin Bolte, Yao-Hung~Hubert Tsai, Kushal Lakhotia, Ruslan Salakhutdinov, and Abdelrahman Mohamed. 2021.
\newblock Hubert: Self-supervised speech representation learning by masked prediction of hidden units.
\newblock \emph{IEEE/ACM Transactions on Audio, Speech, and Language Processing}, 29:3451--3460.

\bibitem[{Huang and Chen(2020)}]{huang2020learning}
Chao-Wei Huang and Yun-Nung Chen. 2020.
\newblock Learning asr-robust contextualized embeddings for spoken language understanding.
\newblock In \emph{ICASSP 2020-2020 IEEE International Conference on Acoustics, Speech and Signal Processing (ICASSP)}, pages 8009--8013. IEEE.

\bibitem[{Kim et~al.(2022)Kim, Yoon, and Jung}]{kim2022improved}
June-Woo Kim, Hyekyung Yoon, and Ho-Young Jung. 2022.
\newblock Improved spoken language representation for intent understanding in a task-oriented dialogue system.
\newblock \emph{Sensors}, 22(4):1509.

\bibitem[{Kim et~al.(2021)Kim, Kim, Shin, and Lee}]{kim2021two}
Seongbin Kim, Gyuwan Kim, Seongjin Shin, and Sangmin Lee. 2021.
\newblock Two-stage textual knowledge distillation for end-to-end spoken language understanding.
\newblock In \emph{ICASSP 2021-2021 IEEE International Conference on Acoustics, Speech and Signal Processing (ICASSP)}, pages 7463--7467. IEEE.

\bibitem[{Lai et~al.(2020)Lai, Cao, Bodapati, and Li}]{lai2020towards}
Cheng-I Lai, Jin Cao, Sravan Bodapati, and Shang-Wen Li. 2020.
\newblock Towards semi-supervised semantics understanding from speech.
\newblock \emph{arXiv preprint arXiv:2011.06195}.

\bibitem[{Li et~al.(2023)Li, Li, Savarese, and Hoi}]{li2023blip}
Junnan Li, Dongxu Li, Silvio Savarese, and Steven Hoi. 2023.
\newblock Blip-2: Bootstrapping language-image pre-training with frozen image encoders and large language models.
\newblock In \emph{International conference on machine learning}, pages 19730--19742. PMLR.

\bibitem[{Lin et~al.(2023)Lin, Zhu, Ye, Ning, Jin, and Yuan}]{lin2023video}
Bin Lin, Bin Zhu, Yang Ye, Munan Ning, Peng Jin, and Li~Yuan. 2023.
\newblock Video-llava: Learning united visual representation by alignment before projection.
\newblock \emph{arXiv preprint arXiv:2311.10122}.

\bibitem[{Liu and Lane(2016)}]{liu2016attention}
Bing Liu and Ian Lane. 2016.
\newblock Attention-based recurrent neural network models for joint intent detection and slot filling.
\newblock \emph{Interspeech 2016}.

\bibitem[{Liu et~al.(2019)Liu, Ott, Goyal, Du, Joshi, Chen, Levy, Lewis, Zettlemoyer, and Stoyanov}]{liu2019roberta}
Yinhan Liu, Myle Ott, Naman Goyal, Jingfei Du, Mandar Joshi, Danqi Chen, Omer Levy, Mike Lewis, Luke Zettlemoyer, and Veselin Stoyanov. 2019.
\newblock Roberta: A robustly optimized bert pretraining approach.
\newblock \emph{arXiv preprint arXiv:1907.11692}.

\bibitem[{Lugosch et~al.(2021)Lugosch, Papreja, Ravanelli, Heba, and Parcollet}]{lugosch2021timers}
Loren Lugosch, Piyush Papreja, Mirco Ravanelli, Abdelwahab Heba, and Titouan Parcollet. 2021.
\newblock Timers and such: A practical benchmark for spoken language understanding with numbers.
\newblock In \emph{35th Conference on Neural Information Processing Systems (NeurIPS 2021) Track on Datasets and Benchmarks.}, pages 1--11.

\bibitem[{Lugosch et~al.(2019)Lugosch, Ravanelli, Ignoto, Tomar, and Bengio}]{lugosch2019speech}
Loren Lugosch, Mirco Ravanelli, Patrick Ignoto, Vikrant~Singh Tomar, and Yoshua Bengio. 2019.
\newblock Speech model pre-training for end-to-end spoken language understanding.
\newblock \emph{arXiv preprint arXiv:1904.03670}.

\bibitem[{Oord et~al.(2018)Oord, Li, and Vinyals}]{oord2018representation}
Aaron van~den Oord, Yazhe Li, and Oriol Vinyals. 2018.
\newblock Representation learning with contrastive predictive coding.
\newblock \emph{arXiv preprint arXiv:1807.03748}.

\bibitem[{Ouyang et~al.(2022)Ouyang, Wu, Jiang, Almeida, Wainwright, Mishkin, Zhang, Agarwal, Slama, Ray et~al.}]{ouyang2022training}
Long Ouyang, Jeffrey Wu, Xu~Jiang, Diogo Almeida, Carroll Wainwright, Pamela Mishkin, Chong Zhang, Sandhini Agarwal, Katarina Slama, Alex Ray, et~al. 2022.
\newblock Training language models to follow instructions with human feedback.
\newblock \emph{Advances in neural information processing systems}, 35:27730--27744.

\bibitem[{Qian et~al.(2021)Qian, Bianv, Shi, Kanda, Shen, Xiao, and Zeng}]{qian2021speech}
Yao Qian, Ximo Bianv, Yu~Shi, Naoyuki Kanda, Leo Shen, Zhen Xiao, and Michael Zeng. 2021.
\newblock Speech-language pre-training for end-to-end spoken language understanding.
\newblock In \emph{ICASSP 2021-2021 IEEE International Conference on Acoustics, Speech and Signal Processing (ICASSP)}, pages 7458--7462. IEEE.

\bibitem[{Qin et~al.(2021)Qin, Liu, Che, Kang, Zhao, and Liu}]{qin2021co}
Libo Qin, Tailu Liu, Wanxiang Che, Bingbing Kang, Sendong Zhao, and Ting Liu. 2021.
\newblock A co-interactive transformer for joint slot filling and intent detection.
\newblock In \emph{ICASSP 2021-2021 IEEE International Conference on Acoustics, Speech and Signal Processing (ICASSP)}, pages 8193--8197. IEEE.

\bibitem[{Radford et~al.(2019)Radford, Wu, Child, Luan, Amodei, Sutskever et~al.}]{radford2019language}
Alec Radford, Jeffrey Wu, Rewon Child, David Luan, Dario Amodei, Ilya Sutskever, et~al. 2019.
\newblock Language models are unsupervised multitask learners.
\newblock \emph{OpenAI blog}, 1(8):9.

\bibitem[{Ruan et~al.(2020)Ruan, Nechaev, Chen, Su, and Kiss}]{ruan2020towards}
Weitong Ruan, Yaroslav Nechaev, Luoxin Chen, Chengwei Su, and Imre Kiss. 2020.
\newblock Towards an asr error robust spoken language understanding system.

\bibitem[{Saxon et~al.(2021)Saxon, Choudhary, McKenna, and Mouchtaris}]{saxon21_interspeech}
Michael Saxon, Samridhi Choudhary, Joseph~P. McKenna, and Athanasios Mouchtaris. 2021.
\newblock \href {https://doi.org/10.21437/Interspeech.2021-1826} {{End-to-End Spoken Language Understanding for Generalized Voice Assistants}}.
\newblock In \emph{Proc. Interspeech 2021}, pages 4738--4742.

\bibitem[{Seo et~al.(2022)Seo, Kwak, and Lee}]{seo2022integration}
Seunghyun Seo, Donghyun Kwak, and Bowon Lee. 2022.
\newblock Integration of pre-trained networks with continuous token interface for end-to-end spoken language understanding.
\newblock In \emph{ICASSP 2022-2022 IEEE International Conference on Acoustics, Speech and Signal Processing (ICASSP)}, pages 7152--7156. IEEE.

\bibitem[{Serdyuk et~al.(2018)Serdyuk, Wang, Fuegen, Kumar, Liu, and Bengio}]{serdyuk2018towards}
Dmitriy Serdyuk, Yongqiang Wang, Christian Fuegen, Anuj Kumar, Baiyang Liu, and Yoshua Bengio. 2018.
\newblock Towards end-to-end spoken language understanding.
\newblock In \emph{2018 IEEE International Conference on Acoustics, Speech and Signal Processing (ICASSP)}, pages 5754--5758. IEEE.

\bibitem[{Su et~al.(2022)Su, Liu, Meng, Lan, Shu, Shareghi, and Collier}]{su2022tacl}
Yixuan Su, Fangyu Liu, Zaiqiao Meng, Tian Lan, Lei Shu, Ehsan Shareghi, and Nigel Collier. 2022.
\newblock Tacl: Improving bert pre-training with token-aware contrastive learning.
\newblock In \emph{Findings of the Association for Computational Linguistics: NAACL 2022}, pages 2497--2507.

\bibitem[{Touvron et~al.(2023)Touvron, Lavril, Izacard, Martinet, Lachaux, Lacroix, Rozi{\`e}re, Goyal, Hambro, Azhar et~al.}]{touvron2023llama}
Hugo Touvron, Thibaut Lavril, Gautier Izacard, Xavier Martinet, Marie-Anne Lachaux, Timoth{\'e}e Lacroix, Baptiste Rozi{\`e}re, Naman Goyal, Eric Hambro, Faisal Azhar, et~al. 2023.
\newblock Llama: Open and efficient foundation language models.
\newblock \emph{arXiv preprint arXiv:2302.13971}.

\bibitem[{Wang and Isola(2020)}]{wang2020understanding}
Tongzhou Wang and Phillip Isola. 2020.
\newblock Understanding contrastive representation learning through alignment and uniformity on the hypersphere.
\newblock In \emph{International Conference on Machine Learning}, pages 9929--9939. PMLR.

\bibitem[{Wang et~al.(2024)Wang, Chen, Chen, Wu, Zhu, Zeng, Luo, Lu, Zhou, Qiao et~al.}]{wang2024visionllm}
Wenhai Wang, Zhe Chen, Xiaokang Chen, Jiannan Wu, Xizhou Zhu, Gang Zeng, Ping Luo, Tong Lu, Jie Zhou, Yu~Qiao, et~al. 2024.
\newblock Visionllm: Large language model is also an open-ended decoder for vision-centric tasks.
\newblock \emph{Advances in Neural Information Processing Systems}, 36.

\bibitem[{Zhao et~al.(2023)Zhao, Wang, Zhang, Fu, Do, Agarwal, Lee, and Sun}]{zhao2023antgpt}
Qi~Zhao, Shijie Wang, Ce~Zhang, Changcheng Fu, Minh~Quan Do, Nakul Agarwal, Kwonjoon Lee, and Chen Sun. 2023.
\newblock Antgpt: Can large language models help long-term action anticipation from videos?
\newblock In \emph{The Twelfth International Conference on Learning Representations}.

\end{thebibliography}
\end{document}